\documentclass[letterpaper, 10 pt, conference]{ieeeconf}  

\IEEEoverridecommandlockouts                              

\usepackage[tracking=true]{microtype}
\usepackage{amsmath} 
\usepackage{amssymb}  
\usepackage{graphicx}
\usepackage{epsfig} 
\usepackage{mathptmx} 
\usepackage{times} 
\usepackage{amsmath} 
\usepackage{amssymb}  
\usepackage{xspace}
\usepackage{url}

\usepackage[dvipsnames]{xcolor}
\usepackage[font={footnotesize}]{caption}
\usepackage{booktabs} 
\usepackage{tikz}
\usetikzlibrary{positioning,shapes.geometric,arrows,fit,shapes.symbols,svg.path,tikzmark,calc}
\usepackage{textcomp}
\usepackage{listings}
\usepackage{subcaption}
\pgfdeclarelayer{background}
\pgfdeclarelayer{foreground}
\pgfsetlayers{background,main,foreground}
\usepackage{siunitx}
\usepackage{adjustbox}
\usepackage{array}
\usepackage{multirow}

\usepackage{siunitx}

\let\labelindent\relax 

\usepackage[inline]{enumitem} 

\usepackage[backend=biber,
            url=false,
            isbn=false,
            doi=false,
            backref=false,
            style=ieee,
            natbib=true,
            mincitenames=1,
            maxcitenames=1,
            citestyle=numeric-comp,
            sorting=none,
            block=none]{biblatex}
\renewcommand{\bibfont}{\small}
\newcommand{\algname}{FogROS2-PLR\xspace}
\newcommand{\sgc}{FogROS2-SGC\xspace}

\title{\LARGE \bf
FogROS2-PLR: Probabilistic Latency-Reliability For Cloud Robotics 
}

\author{$^\dagger$Kaiyuan Chen$^{{1,2}}$, Nan Tian$^{2}$, Christian Juette$^{2}$, Tianshuang Qiu$^{{1,2}}$,  \\ Liu Ren$^{2}$, John Kubiatowicz$^{1}$, and Ken Goldberg$^{1, 3}$
\thanks{$^{1}$Department of Electrical Engineering and Computer Science}%
\thanks{$^{2}$Robert Bosch Research and Technology Center North America, Sunnyvale, CA, USA}%
\thanks{$^{3}$Department of Industrial Engineering and Operations Research}%
\thanks{$^{1,3}$University of California, Berkeley, CA, USA }%
\thanks{$^\dagger$For correspondence and questions: {kych@berkeley.edu}}
}

\begin{document}

\maketitle
\thispagestyle{empty}
\pagestyle{empty}

\begin{abstract}
    Cloud robotics enables robots to offload computationally intensive tasks to cloud servers for performance, cost, and ease of management. 
However, the network and cloud computing infrastructure are not designed for reliable timing guarantees, due to fluctuating Quality-of-Service (QoS).
In this work, we formulate an impossibility triangle theorem for: \underline{L}atency reliability, \underline{S}ingleton server, and \underline{C}ommodity hardware. The LSC theorem suggests that providing replicated servers with uncorrelated failures can exponentially reduce the probability of missing a deadline.
We present FogROS2-Probabilistic Latency Reliability (PLR) that uses multiple independent network interfaces to send requests to replicated cloud servers and uses the first response back. 
We design routing mechanisms to discover, connect, and route through non-default network interfaces on robots. 
\algname optimizes the selection of interfaces to servers to minimize the probability of missing a deadline. 
We conduct a cloud-connected driving experiment with two 5G service providers, demonstrating \algname effectively provides smooth service quality even if one of the service providers experiences low coverage and base station handover. 
We use 99 Percentile (P99) latency to evaluate anomalous long-tail latency behavior.
In one experiment, \algname improves P99 latency by up to 3.7x compared to using one service provider.
We deploy \algname on a physical Stretch 3 robot performing an indoor human-tracking task. 
Even in a fully covered Wi-Fi and 5G environment, \algname improves the responsiveness of the robot reducing mean latency by 36\% and P99 latency by 33\%. 
Code and supplementary can be found on website\footnote{\url{https://github.com/data-capsule/rt-fogros2/tree/udp}}. 
\end{abstract}

\section{Introduction}

The complexity of robotic algorithms~\cite{ichnowski2020fog,tanwani2019fog} and models~\cite{wang2021nerf,kirillov2023segment,Rashid24lifelonglerf} often surpasses the computing capabilities of onboard hardware of robots. 
State-of-the-art Large Language Models  (LLM)~\cite{touvron2023llama,achiam2023gpt4}, Vision Language Models (VLM)~\cite{2023GPT4VisionSC, 2023gemini, liu2024visual, alayrac2022flamingo,driess2023palme} and Vision-Language-Action (VLA)~\cite{octo_2023, open_x_embodiment_rt_x_2023, brohan2023rt1, brohan2023rt2,  kim24openvla, khazatsky2024droid} are large and almost always hosted in cloud environment.
FogROS2~\cite{ichnowski2022fogros} is an open-source Cloud Robotics framework that deploys unmodified compute-intensive algorithms in Robot Operating System 2 (ROS2).
However, Cloud robotics infrastructures are typically implemented to prioritize cost and resource efficiency, often sharing network or compute infrastructure among multiple robots and heterogeneous workloads.
However, as network quality of service (QoS) between robots and the cloud varies, we show providing \emph{Latency-Reliable} cloud robotics services on \emph{Commodity}\footnote{We borrow the term `commodity' from The Google File System~\cite{ghemawat2003google}. The word is widely used in network research~\cite{egi2008towards,sonchack2018turboflow,lu2012using} to describe cheap, latency unreliable and error-prone compute, network and storage infrastructure} infrastructure is impossible with a single cloud server setup.
We introduce \algname that simultaneously uses independent robot network interfaces and independent cloud servers (Figure \ref{fig:intro:use_case}). 
\algname enhance performance against variable network QoS and infrastructure unavailability and improve the responsiveness of cloud robotics despite the following limitations:

\begin{figure}
    \centering
    \includegraphics[width=\linewidth]{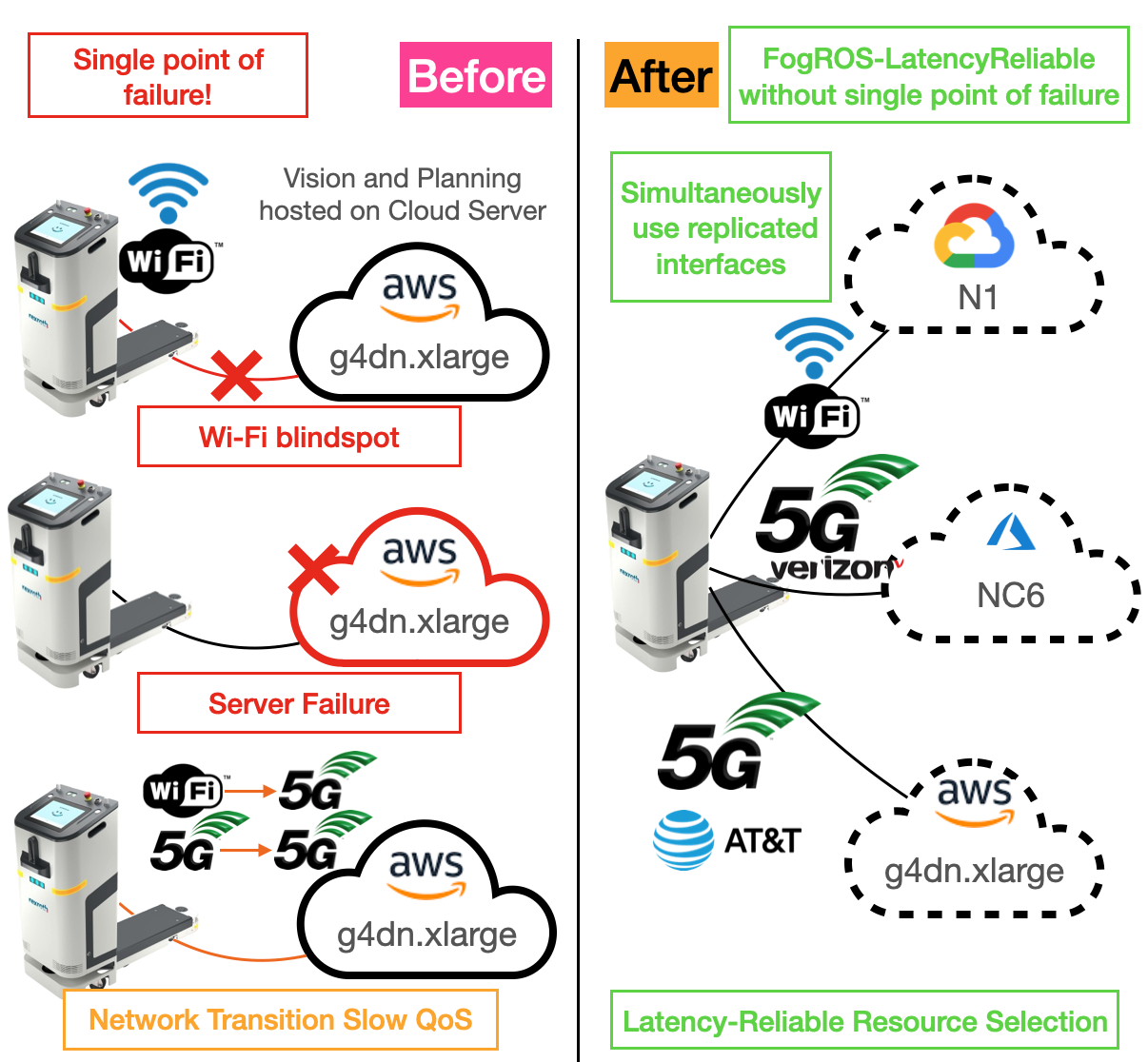}
    \caption{\textbf{\algname Use Case.} A mobile robot in a warehouse connects to the cloud for vision, planning, and coordination. A smooth connection is required for safety and responsiveness. 
    \textbf{(Left)} Conventional cloud robotics is subject to a single point of failure. In the top and middle, network or server failure leads to a complete breakdown of the system. At the bottom, transition to an alternative network or server at slowdown leads to QoS degradation. 
    \textbf{(Right)} Instead, \algname provides a fault-tolerant solution that deploys unmodified ROS2 applications to multiple low-cost cloud servers, making cloud-robotics applications resilient to individual server termination and network slowdowns.
    }
    \label{fig:intro:use_case}
\end{figure}

\begin{enumerate}[label=(\Alph*),itemindent=3em,leftmargin=0pt]

\item \textit{Latency-Unreliable Computational Infrastructure}:
The cloud allows for flexible use of computational resources. For instance, systems can be oversubscribed by allocating fewer resources than the total required by all robots, assuming that not all robots demand resources simultaneously. This approach enhances resource utilization, but if too many robots access resources simultaneously, it can lead to latency degradation or even failures.

\item \textit{Latency-Unreliable Network Infrastructure}: 
The network infrastructure connecting robots to the cloud can experience varying levels of reliability. In cloud robotics, the network's quality of service (QoS) is crucial for maintaining consistent communication and performance. However, fluctuations in network bandwidth or latency can occur due to factors such as network congestion, physical distance, and network outages.

\item \textit{System Dynamism}: 
Latency fluctuations can sometimes arise not from the infrastructure itself but from the underlying mechanisms. For instance, in a 5G network hosted by a base station, handovers are required to maintain connectivity as a vehicle moves from one cell to another. In addition, a mobile robot in a warehouse may need to switch between an indoor Wi-Fi connection and a 5G cellular connection. The operating system must detect signal reduction or link breakdown and then initiate a switch to an alternative network. This process can take anywhere from 100 milliseconds to 10 seconds~\cite{tan2018supporting}.
In both scenarios, frequent network switching can cause service interruptions, while delayed switching can result in slow quality of service.
\end{enumerate}

Specially designed real-time operating systems~\cite{ren2020fine, abderrahim2019efficient, shao2022edge} and network communication protocols~\cite{shukla2023improving, 10.1155/2020/8517372,8792078,9197657} rely on dedicated resources to  mitigate latency unreliability. 
However, deploying such systems is expensive, and requires overprovisioning network and compute resources.
In addition, in public networks, achieving dedicated resource allocation might not even be possible due to policy decisions, such as those surrounding net neutrality. Public networks are designed to treat all data equally, which limits the ability to prioritize certain types of traffic or allocate specific resources to a single application. This restriction makes it challenging to implement real-time systems that rely on guaranteed low-latency performance in such environments.
 

Alternatively, we formulate probabilistic latency-reliable cloud robotics operating on unreliable commodity infrastructure. 
We find that providing replicated resources with uncorrelated failures can reduce the failure probability exponentially. 
We propose \algname, a cloud robotics framework that uses multiple independent networks for the robot and failure-independent compute resources to achieve probabilistic latency reliability on commodity cloud infrastructure. 
\algname discovers, connects, and simultaneously routes messages through multiple non-default network interfaces on robots to identical services deployed in multiple cloud data centers and uses the first response received from the replicated services. This model significantly increases the probability of getting timely responses as long as at least one replica and the network remain operational and responsive.
\algname optimizes the selection of interfaces to cloud-deployed robotics algorithms by minimizing the probability of deadline misses. 

We evaluate \algname with a cloud-operated driving experiment with two 5G service providers, demonstrating \algname effectively provides smooth quality of service even if one of the service providers experiences low coverage and base station handover. In the experiment, \algname improves anomalous long tail P99 (99 Percentile) latency by up to 3.7x. 
We deploy \algname on a physical Stretch 3 robot with an indoor human-tracking task. 
Even in a fully covered Wi-Fi and 5G environment, \algname the responsiveness of the robot by reducing 36\% of mean latency and 33\% of anomalous long-tail latency. 

This paper makes five contributions: 
(1) Formulation of Probabilistic Latency-Reliability for cloud robotics;
(2) LSC Theorem that latency reliability, singleton deployment, and commodity infrastructure cannot co-exist;
(3) \algname, a cloud robotics framework that uses independent network interfaces and cloud services for latency reliability;
(4) Algorithm to determine the optimal choice of network interface-server combination to maximize latency reliability;
(5) Evaluation of \algname on human tracking and autonomous driving applications.

\section{Related Work}

\paragraph*{Cloud and Fog Robotics}
The use of cloud computing resources for robots conceptualized as cloud robotics~\cite{kehoe2015survey}, has become increasingly relevant
as large models. These computationally demanding models/ algorithms have a wide range of applications in robotics, such as visual perception ~\cite{wang2021nerf,kirillov2023segment,kerr2023lerf,Rashid24lifelonglerf}).
Following the Fog Computing paradigm~\cite{bonomi2012fog}, Fog Robotics~\cite{gudi2017fog} utilizes edge resources to improve performance, enabling the viability of cloud computing for a multitude of robotics applications~\cite{tanwani2019fog,ichnowski2020fog,tian2019fog,gudi2018fog}.
FogROS2~\cite{chen2021fogros} is a cloud robotics framework officially supported by ROS2~\cite{macenski2022robot}.
The extension of FogROS2, \sgc~\cite{chen2023sgc} enables secure communication between distributed ROS2 robot nodes. of this work has addressed
the questions of connectivity, latency, and cost.

\paragraph*{Latency Sensitive Cloud and Fog Infrastructure} Edge or Cloud real-time systems~\cite{ren2020fine, abderrahim2019efficient} on commodity hardware typically assume the networking has reliable latency. For example, Edge-RT\cite{shao2022edge} makes the best effort to guarantee the deadline after the packet is received.  
However, providing low-latency networking in wireless network settings for automation and robotics requires special hardware such as radio frequency, wireless channels, and environment assumptions \cite{shukla2023improving, 10.1155/2020/8517372,8792078,9197657}. 
\algname recognizes that deploying such systems can be expensive and aims for a practical and general latency reliable cloud robotics framework.

\paragraph*{Latency Sensitive Cloud Robotics via Redundancy}
Existing work uses \emph{redundancy} by duplicated machines to prevent failures of one machine.
\citet{schafhalter2023cloudav} improves the responsiveness of autonomous vehicles by performing operations on both vehicle and cloud. 
FogROS2-LS (Latency Sensitive)~\cite{chen2024fogrosls} uses replicated communication and compute resources so that a robot can flexibly connect to using one of many servers, but the system takes time to discover and recover from faults by switching to another server that meets latency requirements.
\citet{wu2018visual} statistically models and detects anomalous time series events.
Prior work also designed applications to fault tolerance environments with spot VMs, such as web services~\cite{alieldin2019spotweb} and deep learning~\cite{lee2017deepspotcloud}.

\section{Impossibility Triangle of Probabilistic Latency-Reliable Cloud Robotics}
We consider a closed-loop, deterministic and stateless cloud robotics task. The robot \(r\) connects independently with the independent server(s) \(S\) that hosts the cloud robotics service.
All together they form a network graph \(G = (V,\ E)\), where \(V\) is the set of nodes (including \(r\), servers \({S}\)
, and \(E\) is the set of edges representing the communication links. One can deploy multiple servers, so we use $\mathcal{P}_{rS}$ to denote all the paths that the robot connects to possible servers, and 
a robot $r$ can connect to a cloud server $s \in S$ with a bidirectional path $p_{rs} \in \mathcal{P}_{rS}$.


The robot sends a request and awaits the server $s$'s response through path $p_{r \rightarrow s}$ with the round-trip latency $L_{p_{rs}}$, which is the sum of the network latency from robot \(r\) to server \(s\) on the path $p_{r \rightarrow s}$ and the processing latency of the server $L_s$.

Given a latency deadline $L_D$, the cloud robotics service is defined as \textbf{latency reliable} if the probability of latency \(L\) exceeding a threshold \(L_D + \delta\) is less than \(\epsilon\) for all \(\epsilon > 0\) and \(\delta > 0\). Formally,
\[
P(L \leq L_D + \delta) < \epsilon
\]

Typically, Cloud Robotics is a \textbf{singleton deployment} such that there is only one server $s$ that hosts the service ($S = \{s\}$) \underline{and} to server $s$, exactly one path, \(p_{rs}\), is selected to connect \(r\) to \(s\) at any given point in time. In this case, $L = L_{p_{rs}}$.


The robot and cloud are deployed with latency-unreliable \textbf{commodity infrastructure} such that each path $p_{rs} \in \mathcal{P}_{rs}$ from robot $r$ to server $s$ follows a latency probability distribution \(P(L_{{p}_{rs}})\) that for some $\delta_0 > 0$, $\epsilon_0 > 0$,
\begin{equation}
    P(L_{{p}_{rs}} > L_D + \delta_0) \geq \epsilon_0,
    \label{eq:def:commodity_hw}
\end{equation}

With the definitions, we formulate the following theorem:

\textbf{{LSC Theorem}} Among \underline{L}atency Reliability, \underline{S}ingleton deployment, and \underline{C}ommodity infrastructure, a cloud robotics system can have at most two of these three properties.


\begin{figure}
    \centering
    \includegraphics[width=0.8\linewidth]{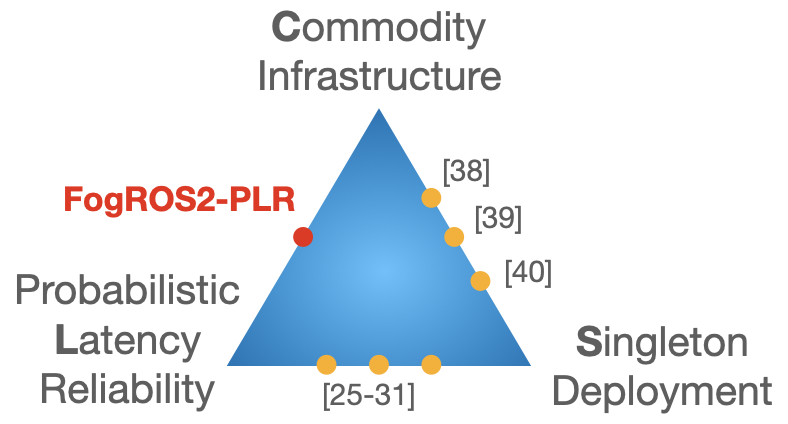}
    \caption{\textbf{Impossibility Triangle of LSC Theorem} Among probabilistic {L}atency Reliability, {S}ingleton deployment, and {C}ommodity infrastructure, a cloud robotics system can have at most two of these three properties. We characterize \algname and its related work on the edges of the impossibility triangle.}
    \label{fig:triangle}
\end{figure}

\textbf{Proof Sketch} We show that assuming any two of three properties implies that the third property cannot be achieved.

\textit{Case 1: Singleton Deployment and Commodity Infrastructure}
With singleton deployment, one path, $p_{rs}$ is selected with latency $L = L_{p_{rs}}$. 
If the infrastructure is latency unreliable, the latency \(L\) on this single path does not meet the reliability requirement in Equation \ref{eq:def:commodity_hw} with $\epsilon_0$ and $\delta_0$.
Thus, latency reliability cannot be guaranteed if the infrastructure is unreliable and there is only one path. This directly contradicts the latency reliability property.

\textit{Case 2: Singleton Deployment and Latency Reliability}
With singleton deployment, there is exactly one path for communication, and achieving latency reliability implies this path must consistently meet the latency requirements. However, if the infrastructure uses commodity hardware it is by definition latency unreliable. Thus the single path fails to meet the latency requirements, making latency reliability impossible. This creates a contradiction and thus cannot use generic infrastructure. 

\textit{Case 3: Latency Reliability and Commodity Infrastructure}
Commodity infrastructure always have a non-negligible probability in the latency requirement in Equation \ref{eq:def:commodity_hw} with $\epsilon_0$ and $\delta_0$.
Therefore, the only way to achieve latency reliability in such an infrastructure is to use more than one independent network and server.   




\textbf{Theoretical Implication} 
To achieve latency reliability, 
one can choose the (singleton deployment, reliability) edge in the triangle (Fig. \ref{fig:triangle}) to use dedicated specialized infrastructure with real-time guarantees. 
However, deploying such infrastructure can be both resource inefficient and expensive~\cite{ren2020fine, abderrahim2019efficient, shao2022edge, shukla2023improving}.

Alternatively, one can choose (commodity hardware, reliability) edge by providing independent server redundancy $|S| > 1$ and independent network path redundancy that $|\mathcal{P}_{rS}| > 1$ on top of an unreliable infrastructure. 
Suppose the system consists of $n$ possible servers $\{s_i\}_{i=1}^n$ and to server $s_i$, 
$m_i$ independent network paths $\{p_{ij}\}_{j=1}^{m_i} \subseteq \mathcal{P}_{rs}$. 
For a single path \(p_{ij}\) to server \(S_j\), the probability that its latency $L_{p_{ij}} $ exceeds the threshold \(L_D + \delta\) is:
\[
P(L_{p_{ij}} > L_D + \delta) = \epsilon_{ij}.
\]

The independence of paths and servers implies that the total system latency \(L\) is determined by the best combination of path and server. The probability that all combinations exceed \(L_D + \delta\) is:
\[
P(L > L_D + \delta) = \bigcap_{i=1}^m \bigcap_{j=1}^n P\left( L_{p_{ij}} > L_D + \delta \right)
\]
The product decreases exponentially with the number of paths \(m\) and servers \(n\)
$(\epsilon')^{mn} < \epsilon$ for $\epsilon > 0$. 
Specifically, 
\[
\lim_{m \to \infty, n \to \infty} \bigcap_{i=1}^m \bigcap_{j=1}^n \epsilon_{ij} = 0
\]
This shows that increasing the number of independent paths and servers significantly reduces the probability of exceeding the latency threshold.

\textbf{Practical Implication}
The assumption of independence is important because it allows us to multiply the probabilities of individual paths to determine the overall probability of meeting latency requirements.
However, in practice, achieving the independence is challenging because users typically cannot control the routing of packets once they enter the Internet service providers infrastructure. 
To approximate independence, we can use multiple network interfaces on the robot, and connect them to different and independent Internet Service Provider (ISP) backbone network. For example, a robot connecting to a Wi-Fi network provided by Comcast is unlikely to experience correlated failures with its connection to a 5G cellular network with AT\&T. 

In summary, we make the following assumptions for practical deployment: 
\begin{enumerate*}
    \item Different network backbones do not experience correlated failures.
    \item Onboard processing overhead of an additional network interface is negligible.
    \item The robots stay within the planned probabilistic reliability distribution.
\end{enumerate*}

\section{\algname System Design}


\begin{figure}
    \centering
    \includegraphics[width=0.9\linewidth]{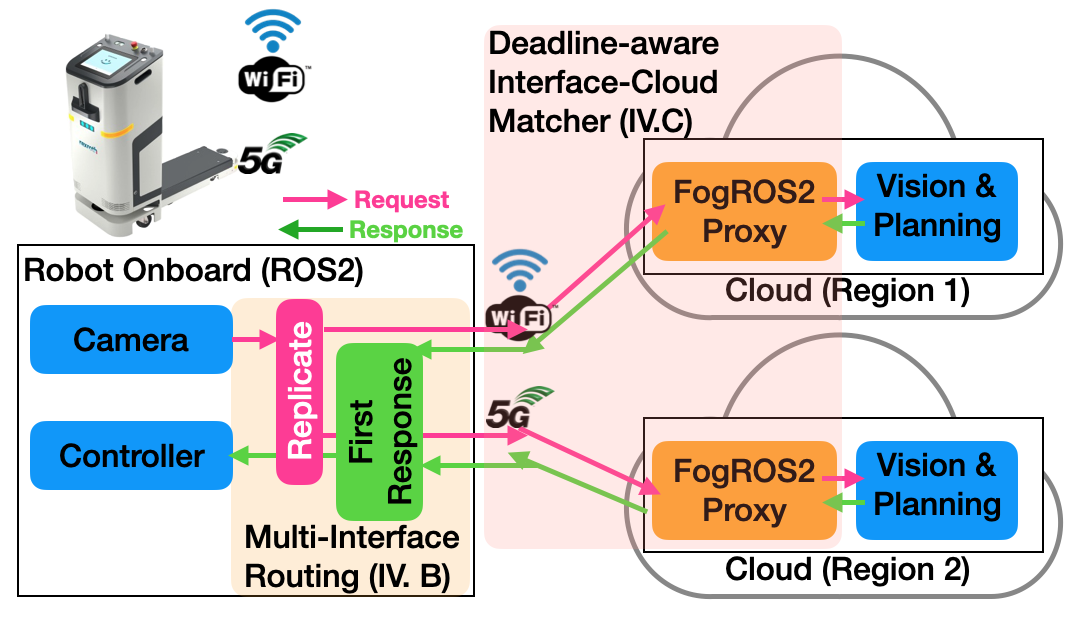}
    \caption{\textbf{System Overview of \algname}  \algname transparently proxies ROS2 communication. It sends requests through multiple network interfaces (such as 5G and Wi-Fi) to replicated Cloud VMs. It uses the first response back to the robot.}
    \label{fig:design:overview}
\end{figure}

Figures \ref{fig:design:overview} show an overview of how \algname achieves latency reliability. 
All the designs of \algname are implemented within a proxy that sits between ROS2 and our custom network protocol. This proxy allows unmodified ROS2 applications to seamlessly integrate with our framework.
\algname replicates requests over multiple independent network interfaces to cloud servers at different regions and routes the first response back to the robot. 
Different interface selections and independent regions are used to reduce the correlated failures. 
Section IV.A presents how \algname matches the interface to the cloud server to minimize the probability of deadline miss.  
Section IV.B discusses how \algname discovers and connects with the cloud with multiple network interfaces.

\subsection{Deadline-Aware Interface-Cloud Matcher}

To determine \textit{which} interface should be used to connect to \textit{which} cloud server, \algname 
minimizes the probability of missing the latency deadline by the deadline-aware interface-cloud matcher.

Suppose the system consists of $n$ possible servers $\{s_j\}_{j=1}^n$ and on the robot, there are $m$ interfaces that are able to connect to those servers. 
We use a binary decision variable \(x_{ij}\) denoting if we should use interface $i$ to connect with server $s_j$,
\[
x_{ij} = \begin{cases}
1 & \text{if interface } i \text{ is connected to server } s_j \\
0 & \text{otherwise}
\end{cases}
\]
and the latency $\epsilon_{ij}$ is probability of missing the deadline using interface $i$ to connect with server $s_j$.

We write the objective function as: 
\[
\text{Minimize} \quad \sum_{i=1}^{m} \sum_{j=1}^{n} \epsilon_{ij} \cdot x_{ij}
\]
subject to constraints: 
\begin{enumerate}
    \item Binary decision variables:
    $x_{ij} \in \{0, 1\} \quad \forall i \in \{1, 2, \ldots, m\}, \forall j \in \{1, 2, \ldots, n\}$ 
    \item Each robot interface should connect to exactly one server:
    $\sum_{j=1}^{n} x_{ij} = 1 \quad \forall i \in \{1, 2, \ldots, m\}
    $
    \item The number of connections to each server must not exceed its capacity:
    $\sum_{i=1}^{m} x_{ij} \leq 1 \quad \forall j \in \{1, 2, \ldots, n\}$
\end{enumerate}

This problem can be solved using integer programming~\cite{glpk} to find the optimal assignment of robot interfaces to servers that minimize latency while satisfying the constraints. $L_{ij}$ can be measured by estimating separately with the computational time the compute and network time.
This reduces the overhead of profiling. For example, if the system only uses one class of cloud servers for computation, only the network data needs to be collected by permuting the network interfaces to the cloud servers.

\begin{figure*}
    \centering
    \includegraphics[width=0.75\linewidth]{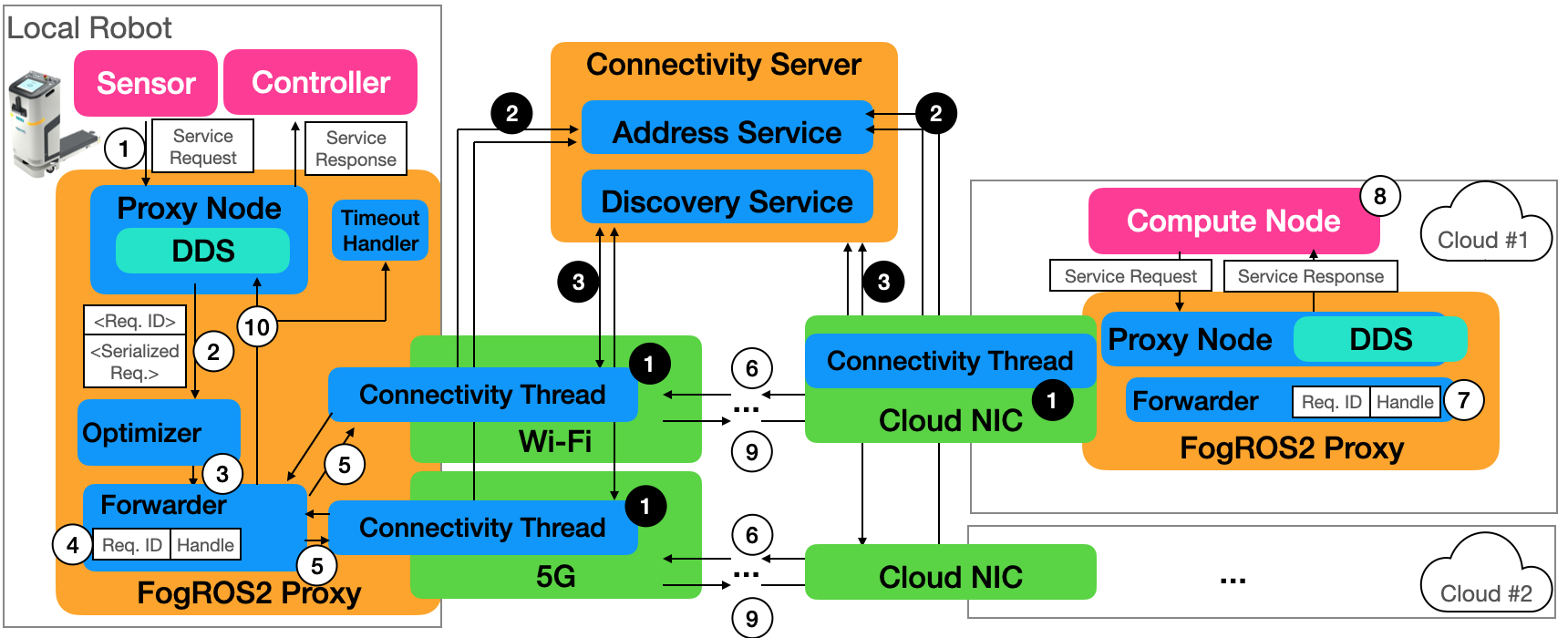}
    \caption{
    \textbf{Workflow Diagram of \algname} On Setup (Black circle), 
    \textbf{(1)} \algname proxy instantiates threads per interface (5G, Wi-Fi, and Cloud Network Interface Card (NIC) in {Green}) per service to handle communication ;
    \textbf{(2)} The thread communicates with a centralized connectivity server. The connectivity server runs STUN protocol to get the public IP address of the given interface; 
    \textbf{(3)} The thread advertises the service-address binding to a centralized discovery service. The discovery service facilitates the cloud and robot to discover each other.
    To handle an incoming request from the robot (White circle), 
    \textbf{(1)} The ROS2 application sends a request to \algname proxy;
    \textbf{(2)} The proxy retrieves a unique request ID from ROS2 Data Distribution Service (DDS) and raw request in bytes; 
    \textbf{(3)} The request goes through optimizer that determines network interface and cloud server mapping;
    \textbf{(4)} The forwarder registers the unique request ID from rmw with a callback for response and a callback for timeout;
    \textbf{(5)} The forwarder replicates and sends the request to the corresponding server specified by the optimizer mapping;
    \textbf{(6)} The message is routed in the failure-independent networks from the robot to the cloud;
    \textbf{(7)} The \algname proxy converts the request to a regular ROS2 service request;
    \textbf{(8)}  The cloud proxy invokes the cloud ROS2 service and gets the response;
    \textbf{(9)} The response is forwarded back the robot;
    \textbf{(10)} The robot proxy uses the first received response from replicated network interfaces and servers, and returns with a regular ROS2 service response to the application. It can optionally invoke a timeout callback if the response is not returned promptly. 
    }
    \label{fig:design:flow}
\end{figure*}

\subsection{Multi-Interface Connectivity} 

Figure \ref{fig:design:flow} shows the workflow of \algname on handling requests with fault tolerance. At the connection setup phase, \algname creates a thread for every available interface available to the robot and creates a socket that forces it to bind to the interfaces even if the interfaces are not default.
Getting the cloud-connectable network addresses from non-default interfaces is challenging, because some local or virtual interfaces, such as Virtual Private Network, are private to the robot. We use a cloud-deployed Session Traversal Utilities for NAT (STUN) protocol~\cite{sredojev2015webrtc} to get the cloud-accessible network IP address. The thread is canceled if it is unable to connect to address STUN service for a period of time. The thread advertises itself to the discovery service; the discovery service runs a custom protocol to exchange the addresses of discovered addresses similar to FogROS2-SGC~\cite{chen2023sgc}. Both address service and discovery service can be held in public Internet or in private network settings.  These services are solely used for establishing connectivity, and no application data passes through the server. Therefore, a server failure does not result in a system failure.

When the robot sends a ROS2 request, \algname's proxy on the robot intercepts it, and extracts its serialized request and a unique identifier from the ROS2 middleware layer (rmw). The proxy then stores this identifier with a handle and replicates the request to the cloud through multiple interfaces. Deadline-Aware Interface-Cloud Matcher constantly updates the mapping of Interface-Cloud.
The cloud proxy executes the corresponding cloud-based ROS2 service, computes the result, and sends the response back to the robot's proxy. The robot's proxy then verifies the response using the unique identifier. If the identifier matches, the request is marked as completed and the response is sent to the robot; otherwise, it is discarded as a duplicate.

\textbf{Multi-Interface Asynchronous Proxy}
While each interface independently sends packets, \algname must ensure efficient internal processing. To achieve this, \algname creates a separate thread for each interface to handle packet processing and potential retransmission in case of packet loss. Maintaining efficiency is crucial; inefficient packet processing compromises the independence property of additional network interfaces: processing on one interface could block others, and the sequence of packet sending would violate the independence assumption.
\algname uses asynchronous operations throughout the packet processing pipeline, allowing the proxy to continue processing packets from other interfaces without delay. Additionally, in \algname, packets are processed with a zero-copy property, meaning that the packet data is not duplicated in memory during processing. This minimizes the overhead associated with using multiple interfaces, preserving efficiency and maintaining the integrity of the independence assumption.

\section{Experiments}

We evaluate the latency reliability of \algname with (1) a simulated robot vision with semantic segmentation with two Wi-Fi interfaces, (2) a cloud operation in a high mobility driving setting with two 5G interfaces, (3) an indoor human tracking with Wi-Fi and 5G. 
We quantify \textit{long-tail} anomalous latency faults with 99 Percentile (P99) latency, the runs with the top 1\,\% latency.

\subsection{Case Study: Cloud Operation with High Mobility}

\begin{figure*}
    \centering
    \includegraphics[width=\linewidth]{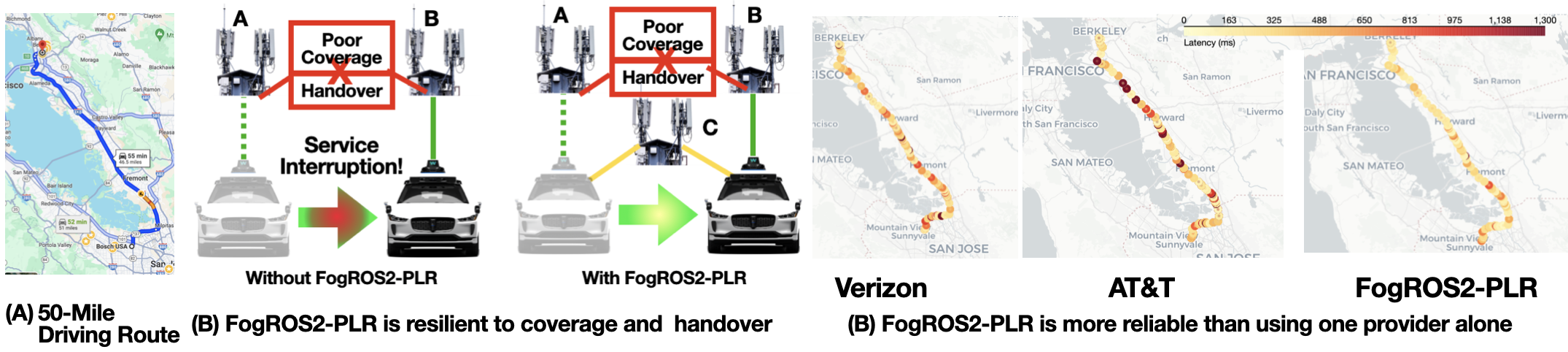}
    \caption{\textbf{Case Study (A) High Mobility Cloud Operation  with \algname}  
    (A) 50-mile Driving route from Sunnyvale, CA to Berkeley, CA with 50 miles. (B) In the mobility test, the car experiences a coverage blind spot of 5G service providers. 5G \emph{handover}: the base station passes the connectivity to the next base station at mobility. Both coverage and handover lead to QoS degradation. \algname prevents such interruption by an independent 5G service provider. (3) While relying on a single service provider leads to QoS fluctuations, \algname demonstrates smooth connectivity by using multiple 5G networks. \algname improves P99 latency of AT\&T and Verizon respectively by 3.7x and 2.4x; it improves mean latency by respectively 2.7x and 1.9x.}
    \label{fig:driving}
\end{figure*}

\textbf{Motivation} 
Autonomous vehicles, such as Waymo, Cruise, and the upcoming Robotaxi use cloud operation to remotely take control in emergencies, such as when a car gets stuck and the onboard algorithms cannot resolve the issue. This cloud operation may involve a teleoperator, a remote data server or a more sophisticated model~\cite{schafhalter_leveraging_2023}. Our study focuses on how to maintain high-quality service in scenarios involving high mobility.

\textbf{Setup} We conducted a 50-mile drive along a standard U.S. highway from Sunnyvale, CA to Berkeley, CA, over a 90-minute period. 
The car's speed was maintained by Tesla's Full Self Driving. During the drive, we streamed 1280x820 resolution images as JPEG from an Intel RealSense 435i camera connected to a laptop. 
The laptop was connected via physical cables to two different 5G hotspots provided by Verizon and AT\&T. We emulated the decision-making process of a cloud operator using object detection with YOLOv8~\cite{redmon2015you}. The end-to-end latency was measured from the moment an image was captured to the time a response was received.

\textbf{Results} Figure \ref{fig:driving} compares \algname with the use of a single service provider. In comparison, AT\&T had higher mean and P99 latency in some extreme cases. \algname effectively leveraged both service providers, offering more reliable latency. \algname improves P99 latency of AT\&T and Verizon respectively by 3.7x (4829.35ms vs 1303.00ms) and 2.4x (3208.04 vs 1303.00). 
It improves mean latency (289.44ms) than AT\&T (791.02ms) and Verizon (550.28ms). We observed a slight failure case near Fremont, CA, where both service providers exhibited higher latency for some requests, likely due to coverage issues relative to the car's speed.

\subsection{Case Study: Human Following Robot}

\begin{figure}
    \centering
    \includegraphics[width=\linewidth]{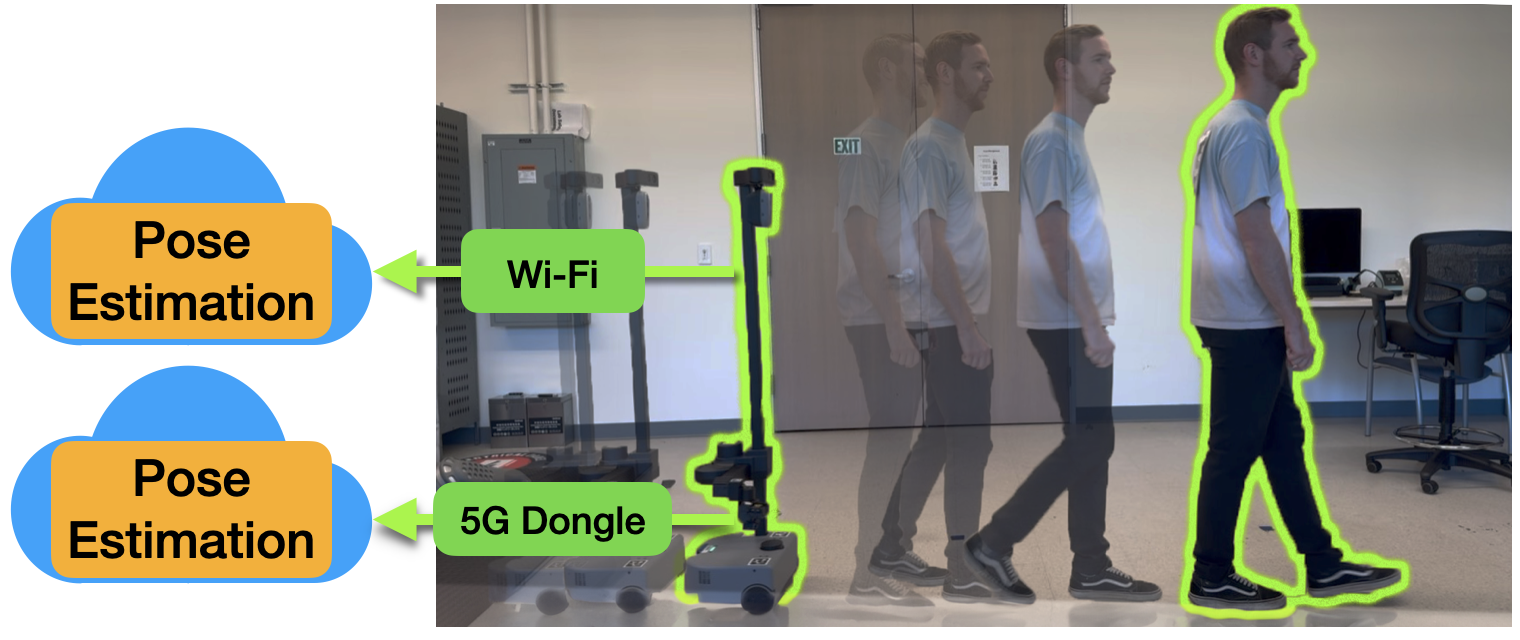}
    \caption{\textbf{Case Study (B) Human Following Robot with \algname in a fully-covered 5G and Wi-Fi environment Setup} We deployed \algname on the Hello Robot Stretch 3 Mobile Manipulator. Stretch Robot connects to two cloud servers running human pose estimation via both 5G and Wi-Fi with \algname.  }
    \label{fig:enter-label}
\end{figure}

\begin{figure}
    \centering
    \includegraphics[width=\linewidth]{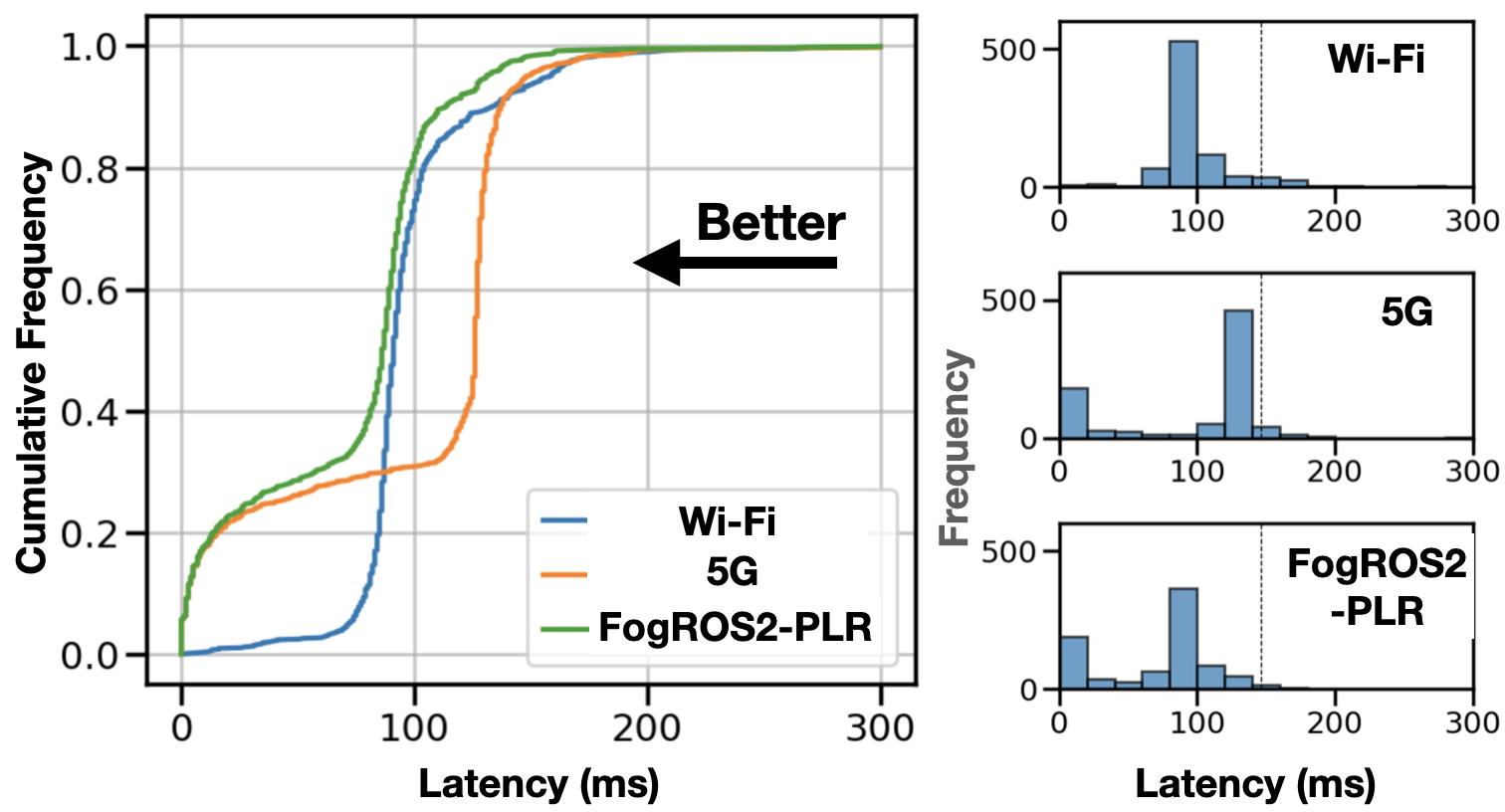}
    \caption{\textbf{Case Study (B) Human Tracking with \algname Results}  (3) Results of \algname. (Left) The comparison of Wi-Fi and 5G latency distributions shows that Wi-Fi generally offers better mean latency, while 5G excels in low-latency scenarios but exhibits less reliable overall latency. \algname effectively combines the strengths of both interfaces, optimizing latency performance.  \algname is able to take advantage of both of the interfaces. (Right) The Figure shows the histogram of latency with a dashed line marking P99 of \algname compared to other options demonstrating that \algname consistently outperforms using either network alone.}
    \label{fig:eval:tracking}
\end{figure}
\textbf{Setup}:
We deployed \algname on the Hello Robot Stretch 3 Mobile Manipulator, running off-the-shelf Ubuntu 22.04 and ROS2 Humble. The mobile manipulator was tasked with following a human along an indoor route around an office. We programmed the tracking and motion based on ~\cite{yolo_tracking}. The robot was connected to both a Wi-Fi dongle with a 5G hotspot and the office's Wi-Fi network.

\textbf{Results} 
Figure \ref{fig:eval:tracking} shows the result of \algname, which improves the P99 long tail latency by 33\% (160.00ms vs 193.52ms) even in an environment with full coverage of Wi-Fi and 5G. \algname can consistently improve the average latency by 36\% (71.39ms vs 97.19ms).

\textbf{Baseline Failure Case} 
Throughout the route, both the 5G and Wi-Fi networks maintained full coverage, despite the varying signaling strength. We attempted to set up a scenario where one network's coverage would drop, requiring the Operating System to switch between Wi-Fi and 5G. Unfortunately, ROS2, using either CycloneDDS (the default DDS before ROS2 Humble) or FastDDS (the default DDS for ROS2 Humble), is unable to switch from the original default interface. This is a known limitation ~\cite{cyclone_interface_fails}.


\section{Conclusion}


In this work, we formulate probabilistic latency reliability on latency-unreliable cloud robotics servers and use failure-independent networks and cloud resources to achieve PLR. The evaluation shows \algname can reduce P99 latency by up to 3.7 times in a realistic driving experiment. 

In future work, we plan to enhance the real-time capabilities of \algname by deploying in a local or edge real-time environment and incorporating a fallback mechanism for handling missed deadlines. Additionally, we will investigate strategies to reduce the cost of using replicated network resources while maintaining high reliability and availability. For example, our earlier work, FogROS2-FT, shows Cloud Spot Virtual Machines can effectively reduce operating costs. In future work, we will explore adaptive scheduling algorithms that can reduce the cost while maintaining PLR.

\newpage 

\renewcommand*{\bibfont}{\footnotesize}
\printbibliography
\end{document}